\documentclass{article} 
\usepackage{arxiv_style,times}

\usepackage{amsmath,amsfonts,bm}

\def\eqref#1{equation~\ref{#1}}

\def\1{\bm{1}}

\DeclareMathAlphabet{\mathsfit}{\encodingdefault}{\sfdefault}{m}{sl}
\SetMathAlphabet{\mathsfit}{bold}{\encodingdefault}{\sfdefault}{bx}{n}

\usepackage{hyperref}
\usepackage{url}
\usepackage[normalem]{ulem}
\usepackage{amssymb}
\usepackage{xcolor}         
\usepackage{graphicx}
\usepackage{float}

\newcommand{\ignore}[1]{{}}

\newcommand{\treepol}{\text{SoftTreeMax}}

\title{\treepol: \\ Policy Gradient with Tree Search}

\author{Gal Dalal\thanks{Equal contribution (random order)},~ Assaf Hallak\footnotemark[1],~ Shie Mannor,~ Gal Chechik \\
NVIDIA Research Israel \\
\texttt{\{gdalal,ahallak,smannor,gchechik\}@nvidia.com}
}

\iclrfinalcopy 
\begin{document}

\maketitle

\begin{abstract}
Policy-gradient methods are widely used for learning control policies. They can be easily distributed to multiple workers and reach state-of-the-art results in many domains. Unfortunately, they exhibit large variance and subsequently suffer from high-sample complexity since they aggregate gradients over entire trajectories. At the other extreme, planning methods, like tree search, optimize the policy using single-step transitions that consider future lookahead. These approaches have been mainly considered for value-based algorithms. Planning-based algorithms require a forward model and are computationally intensive at each step, but are more sample efficient. In this work, we introduce \treepol{}, the first  approach that integrates tree-search into policy gradient. Traditionally, gradients are computed for single state-action pairs. Instead, our tree-based policy structure leverages all gradients at the tree leaves in each environment step. This allows us to reduce the variance of gradients by three orders of magnitude and to benefit from better sample complexity compared with standard policy gradient. On Atari, \treepol{} demonstrates up to 5x better performance in faster run-time compared with distributed PPO.
\end{abstract}

\section{Introduction}

Nowadays, Policy Gradient (PG; \citealt{sutton1999policy}) methods for Reinforcement Learning (RL; \citealt{kaelbling1996reinforcement}) are often the first choice for environments that allow numerous interactions at a fast pace. Their success is attributed to several reasons: they can be easily distributed to multiple workers, require no assumptions on an underlying value function, have both on-policy and off-policy variants, and more.
Despite their popularity, PG algorithms are also notoriously unstable because they compute gradients over entire trajectories \citep{liu2020improved, xu2020improved}. As a result, PG algorithms tend to be highly inefficient in terms of sample complexity. Several solutions were proposed to mitigate this instability. These include baseline subtraction \citep{greensmith2004variance, weaver2001optimal, thomas2017policy, wu2018variance}, anchor-point averaging \citep{papini2018stochastic}, and other variance reduction techniques \citep{papini2018stochastic, zhang2021convergence, shen2019hessian, pham2020hybrid}.

Another family of algorithms that achieved state-of-the-art results in several challenging domains is based on expanding possible future trajectories using tree search (TS; \citealt{coulom2006efficient, silver2009reinforcement}). Prominent examples from this family include UCT \citep{gelly2006exploration}, Dreamer \citep{hafner2019dream}, and MuZero \citep{schrittwieser2020mastering}. TS-based algorithms rely on a forward model (accurate or learned) and substantially reduce sample complexity by offloading the information gain from sampling to planning.
Given a forward modhe main weakness of TS-based methods is that it is difficult to parallelize the learning process because expanding the tree is computationally expensive. However, recent works proposed 
to leverage recent advances in GPU-based simulation and deep model learning for expanding the tree in parallel \citep{dalal2021improve, rosenberg2022planning}.

In this work, we build on top of those recent advances in parallel search and combine PG with TS by introducing a parameterized policy that incorporates tree expansion. Our \treepol{} policy replaces the standard policy-logits of an action with the sum of all logits at the tree leaves that originate from that action. The inference process thus amounts to a fast and efficient breadth-first search followed by vectorized neural network inference, all in the GPU. During training, all gradients at the tree leaves are used for updating the policy, thus significantly mitigating the high variance of PG. 


Our approach addresses a major limitation of TS methods, that their computational complexity grows exponentially with the size of the tree. We reduce the complexity by limiting the width of the tree; we do so by sampling only the most promising nodes at each level. This allows us to efficiently search Atari environments with trees that are $8$-level deep. The fast search mechanism is enabled thanks to a GPU simulator that allows running multiple copies of Atari in parallel \cite{dalton2020accelerating}. We then integrate the resulting \treepol{} into the widely used PPO \citep{schulman2017proximal} and compare it to PPO in its distributed variant. For a fair comparison, we also run the distributed PPO baseline with the parallel GPU emulator by \citep{dalton2020accelerating}. In all tested Atari games, our results outperform the baseline and obtain up to 5x more reward. As we show in Section~\ref{sec: experiments}, the associated gradient variance is three orders of magnitude lower in all games.

\section{Preliminaries}
\label{sec:preliminaries}

We follow the standard notation by \cite{puterman2014markov}. We consider a discounted Markov Decision Process (MDP) $\mathcal{M} = (\mathcal{S}, \mathcal{A}, P, r,\gamma)$, where $\mathcal{S}$ is a finite state space of size $S$, $\mathcal{A}$ is a finite action space of size $A$, $r: \mathcal{S} \times \mathcal{A} \to [0,1]$ is the reward function, $P: \mathcal{S} \times \mathcal{A} \to \Delta_\mathcal{S}$ is the transition function, and $\gamma \in (0,1)$ is the discount factor. 

Let $\pi: \mathcal{S} \to \mathcal{A}$ be a stationary policy, and $V^\pi \in \mathbb{R}^S$ be the value function of $\pi$ defined by $V^\pi(s) = \mathbb{E} \left[ \sum_{t=0}^\infty \gamma^t r(s_t,\pi(s_t)  \mid s_0 = s \right]$.

Our goal is to find the optimal policy $\pi^\star$ such that, for every $s \in \mathcal{S}$,
\[
    V^\star(s)
    =
    V^{\pi^\star}(s)
    =
    \max_{\pi: \mathcal{S} \to \mathcal{A}} V^\pi(s).
\]

When the state space is very large, or continuous, a common practice is to use function approximation to represent the value function, the policy, or even the forward-model. Following the prominent success of deep RL \cite{mnih2015human} in recent years, deep neural networks are used nowadays almost exclusively in practice. Depending on the RL algorithm, a loss function is defined and gradients on the network weights can be calculated.

\subsection{Policy Gradient}
PG schemes seek to maximize the cumulative reward as a function of the parameterized policy $\pi_\theta(a|s)$ by performing gradient steps on $\theta$. For an observed trajectory $\tau=(s_0, a_0, r_0, s_1,\dots,s_T, a_T, r_T,s_{T+1})$, let $r(\tau)$ be the cumulative reward of a trajectory $\tau$, and $\pi_\theta(\tau)$ to be probability of observing trajectory $\tau$ under policy $\pi_\theta$: 
\begin{equation}
    \pi_\theta(\tau) = \Pr(s_0) \cdot \prod_{t=0}^T \pi_\theta(a_t|s_t) \Pr(s_{t+1}, r_t | s_t, a_t).
\end{equation}
Using the likelihood ratio trick, we have
\begin{equation}
    \nabla_\theta \mathbb{E}_{\tau \sim \pi_\theta} \left[ r(\tau) \right] =  \mathbb{E}_{\tau \sim \pi_\theta} \left[ r(\tau) \nabla_\theta \log \pi_\theta(\tau \right] = \mathbb{E}_{\tau \sim \pi_\theta} \left[r(\tau) \sum_{t=0}^T \log \pi_\theta(a_t|s_t) \right].
\end{equation}
Subsequently, multiple trajectories can be sampled to apply stochastic gradient descent on the policy parameters $\theta$ forming the backbone of most PG algorithms.

When the action space is discrete, a commonly used parameterized policy is the soft-max applied over learned weights corresponding to each state-action pair:
\begin{equation}
    \pi_\theta(a|s) \propto \exp \left( w_\theta(s, a) \right),
\end{equation}
where the weights are the output of a neural network with $A$ heads, where the state is the input. Soft-max policies became a canonical part of PG \citep{zhang2021convergence, mei2020global,li2021softmax, schulman2017proximal, haarnoja2018soft}. Some alternatives have been proposed to soft-max \citep{mei2020escaping, miahi2021resmax} with relatively rare usage in practice. Due to its popularity, we focus on a tree extension to the soft-max policy. However, the method proposed here is general and can be applied trivially to other discrete or continuous parameterized policies. 

\subsection{Tree Search}
Planning with a TS is the process of using a forward model to consider possible future trajectories and decide on the best action at the root. One famous such algorithm is Monte-Carlo TS (MCTS; \citealt{browne2012survey}) used in AlphaGo \cite{silver2016mastering} and MuZero \cite{schrittwieser2020mastering}. Other principal algorithms such as Value Iteration, Policy Iteration and DQN were also shown to give an improved performance with a tree search extensions \citep{efroni2019combine, dalal2021improve}. The applicability of these extensions largely benefits from the GPU-based simulator which allows spanning the tree efficiently at each step. This coincides with recently released simulators such as Atari-CuLE \citep{dalton2020accelerating}, IsaacGym \citep{makoviychuk2021isaac} and Brax \cite{freeman2021brax} that support advancing multiple environments simultaneously on the GPU. When such a simulator is out of reach, it is also possible learn a forward model \cite{ ha2018world,kim2020learning,schrittwieser2020mastering}.

\section{\treepol}
We define a new parametric policy, \treepol{}, that operates as follows:
\begin{enumerate}
    \item Given a state $s_0$, expand a TS up to depth $d$ to obtain $H$ trajectories: 
    \begin{equation}
    \{(s^h_0, a^h_0, r^h_0, s^h_1, a^h_1, r^h_1, ..., a^h_{d-1}, r^h_{d-1}, s^h_d, a^h_d  \}_{h=1}^H
    \end{equation}
    The actions can be chosen in any manner, where in our experiments we consider all actions exhaustively and prune under-performing branches so we can reach bigger depths.
    \item Feed all state-action pairs at the leaves into a policy network $w_\theta$ at each branch to obtain the logits $\{ {\sum_{t=0}^{d-1} \gamma^t r^h_t + \gamma^d w_\theta(s^h_d, a^h_d)} \}_{h=1}^H$ that correspond to each action at the root $a_0^h$.
    \item The probability to sample an action $a_0=a$ is a soft-max over all trajectories starting with $a^h_0=a$. Mathematically:
    \begin{equation}\label{eq:pol}
        \pi_\treepol(a|s; \theta) \propto \sum_{h: a^h_0=a} \exp \left[ \beta \left(  \sum_{t=0}^{d-1} \gamma^t r^h_t + \gamma^d w_\theta(s^h_d, a^h_d)  \right) \right],
    \end{equation}
    where $\beta$ is an inverse temperature parameter controlling the exploration intensity.
\end{enumerate}

\textbf{Remark 1.} \treepol{} is a natural planning-based generalization of soft-max: For $d=0,$ it reduces to the standard soft-max. When $d \rightarrow \infty,$ the total weight of a trajectory is its infinite-horizon cumulative discounted reward.

\textbf{Remark 2.} \treepol{} considers the sum of all action values at the leaves, corresponding to Q-function based algorithms. Instead, one could use a single-headed network $w_\theta$ for correspondence with value-function based algorithms.

\textbf{Remark 3.} We add the discounted cumulative reward to the calculated weights. This exploits the additional reward information obtained from expanding the tree. The resulting sum also corresponds to the Boltzmann exploration practice \cite{sutton1999policy} where the exponent mimics the Q-function.

\textbf{Remark 4.} When $\beta \rightarrow 0$ and the same number of trajectories is sampled per action, the policy samples actions uniformly (exploration). When $\beta \rightarrow \infty,$ the policy greedily optimizes for the best trajectory (exploitation). 

\begin{figure}[t]
    \centering
\includegraphics[width=1.0\textwidth]{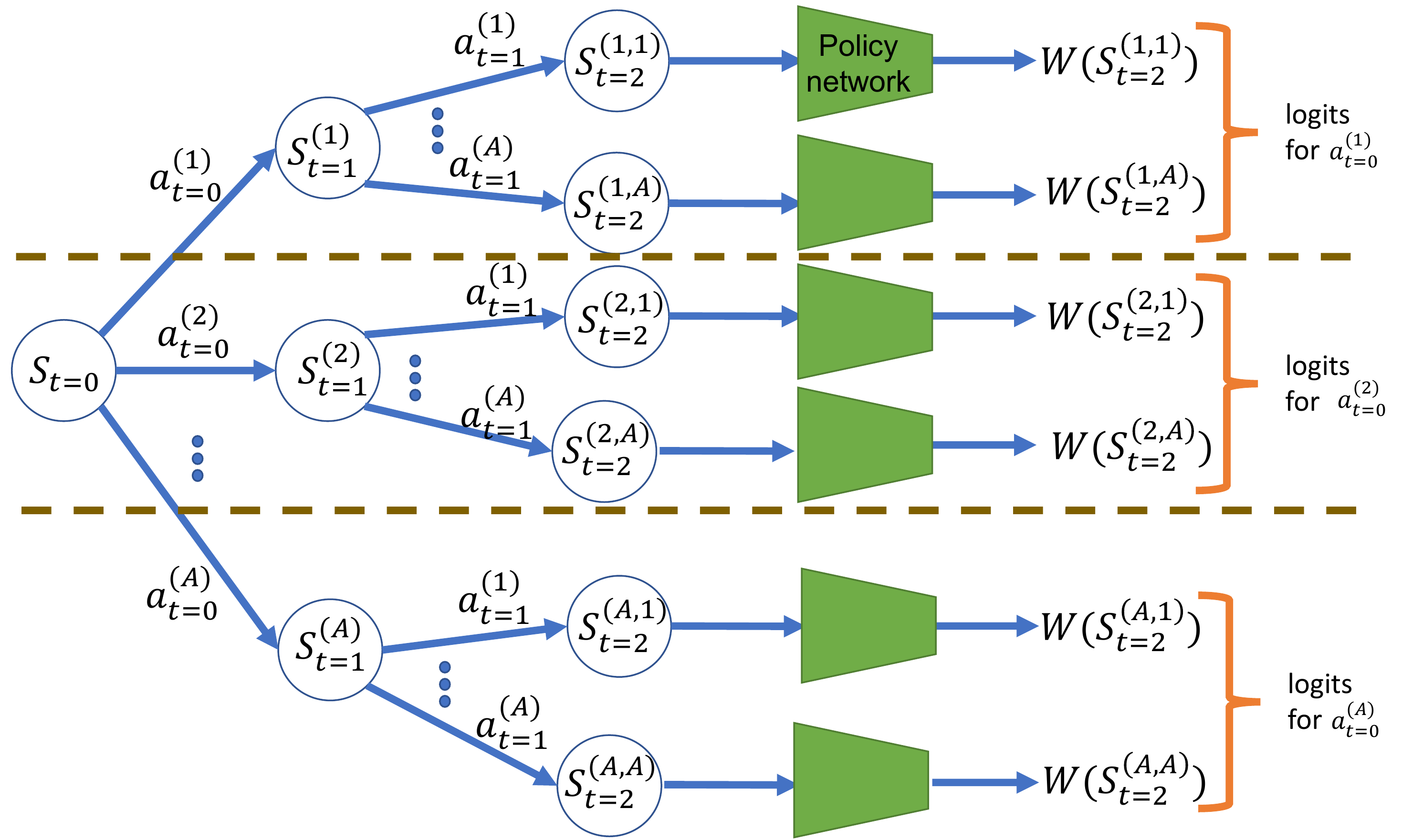}
    \caption{\textbf{\treepol{} policy}. For simplicity, we visualize the exhaustive TS that expands all actions at each state up to depth $d$ ($=2$ here). The leaf state of every trajectory is used as input to the policy network with a different head per action; for simpler visualization we aggregate all actions in the last level. The output is then added to trajectory's cumulative reward as described in \eqref{eq:pol} to form the logit of each action's probability.}  \label{fig:policy_diagram}
\end{figure}

An illustration of the $\treepol$ policy is given in Figure \ref{fig:policy_diagram}. Like other TS methods, \treepol{} requires a forward model that can be used to expand the tree. For an exhaustive search, the complexity of expanding the tree grows exponentially with the tree depth $d.$ However, thanks to our pruning technique, we reduce the complexity to be linear in $d$.

When the gradient $\nabla_\theta \log \pi_\theta$ is calculated at each time step, it updates $w_\theta$ for all state-action pairs in the leaves, similarly to Siamese networks \cite{bertinetto2016fully}. Subsequently, the network is updated even for states that were not observed and actions that were not chosen throughout the simulation trajectory. This effect is essentially a variance reduction technique. As we show in Section~\ref{sec: experiments}, it indeed significantly reduces the noise of the PG process and leads to faster convergence and higher reward.

\section{Experiments}
\label{sec: experiments}
We conduct our experiments on multiple games from the Atari simulation suite \cite{bellemare2013arcade}. As a baseline, we train a PPO \citep{schulman2017proximal} agent with $256$ workers in parallel. In a hyperparameter search, we found this number of workers to be the best in terms of run-time. The environment engine is the highly efficient Atari-CuLE \cite{dalton2020accelerating}, a CUDA-based version of Atari that runs on GPU. Similarly, we use Atari-CuLE for the GPU-based breadth-first TS as done in \cite{dalal2021improve}.
We then train \treepol{} for depths $d=1 \dots 8,$ with a single worker. For brevity, we exclude a few of the depths from the plots. We use five seeds for each experiment. For the implementation, we extend Stable-Baselines3 \cite{raffin2019stable} with all parameters taken as default from the original PPO paper \cite{schulman2017proximal}. We will release the code upon publication. For depths $d \geq 3$, we limited the tree to a maximum width of $1024$ nodes and pruned non-promising trajectories in terms of estimated weights. Since the distributed PPO baseline advances significantly faster in terms of environment steps, for a fair comparison, we ran all experiments for 1 week on the same machine and use the wall-clock time as the x-axis. We use Intel(R) Xeon(R) CPU E5-2698 v4 @ 2.20GHz equipped with one NVIDIA Tesla V100 32GB. 

\begin{figure}[H]
    \centering
\includegraphics[width=1.0\textwidth]{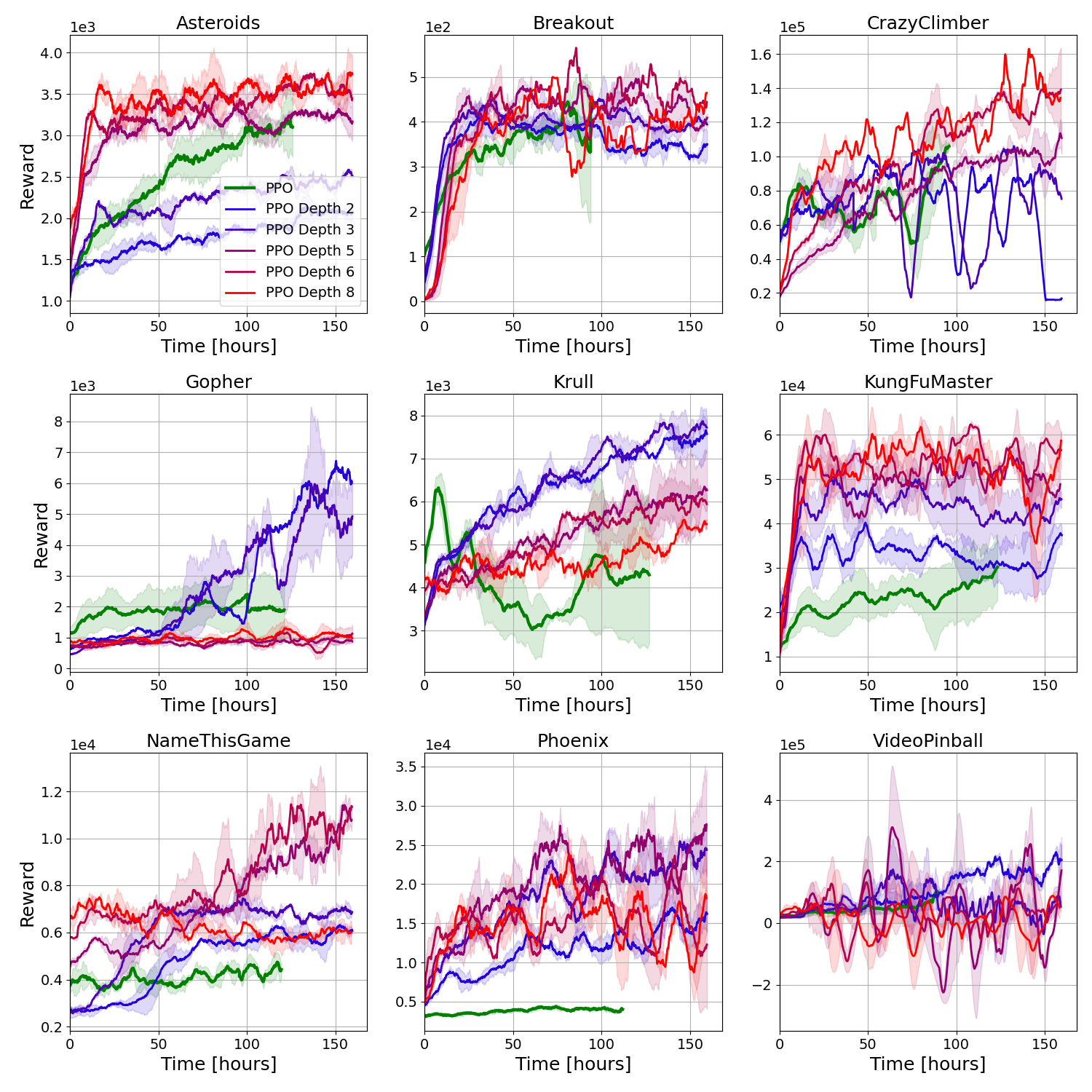}
    \caption{\textbf{Training curves: \treepol{} (single worker) vs PPO ($\bf{256}$ workers).} The plots show average reward and std over five seeds.  The x-axis is the wall-clock time. The maximum time-steps given were 200M, which the standard PPO finished in less than one week of running. }  \label{fig:train_curves}
\end{figure}

We provide the training curves in Figure~\ref{fig:train_curves}. As seen, there is a clear benefit  for \treepol{} over distributed PPO with the standard soft-max policy. In most games, PPO with the \treepol{} policy shows very high sample efficiency: it achieves higher episodic reward even though it observes much less episodes, for the same running time. 

\begin{figure}[H]
    \centering
\includegraphics[width=1.0\textwidth]{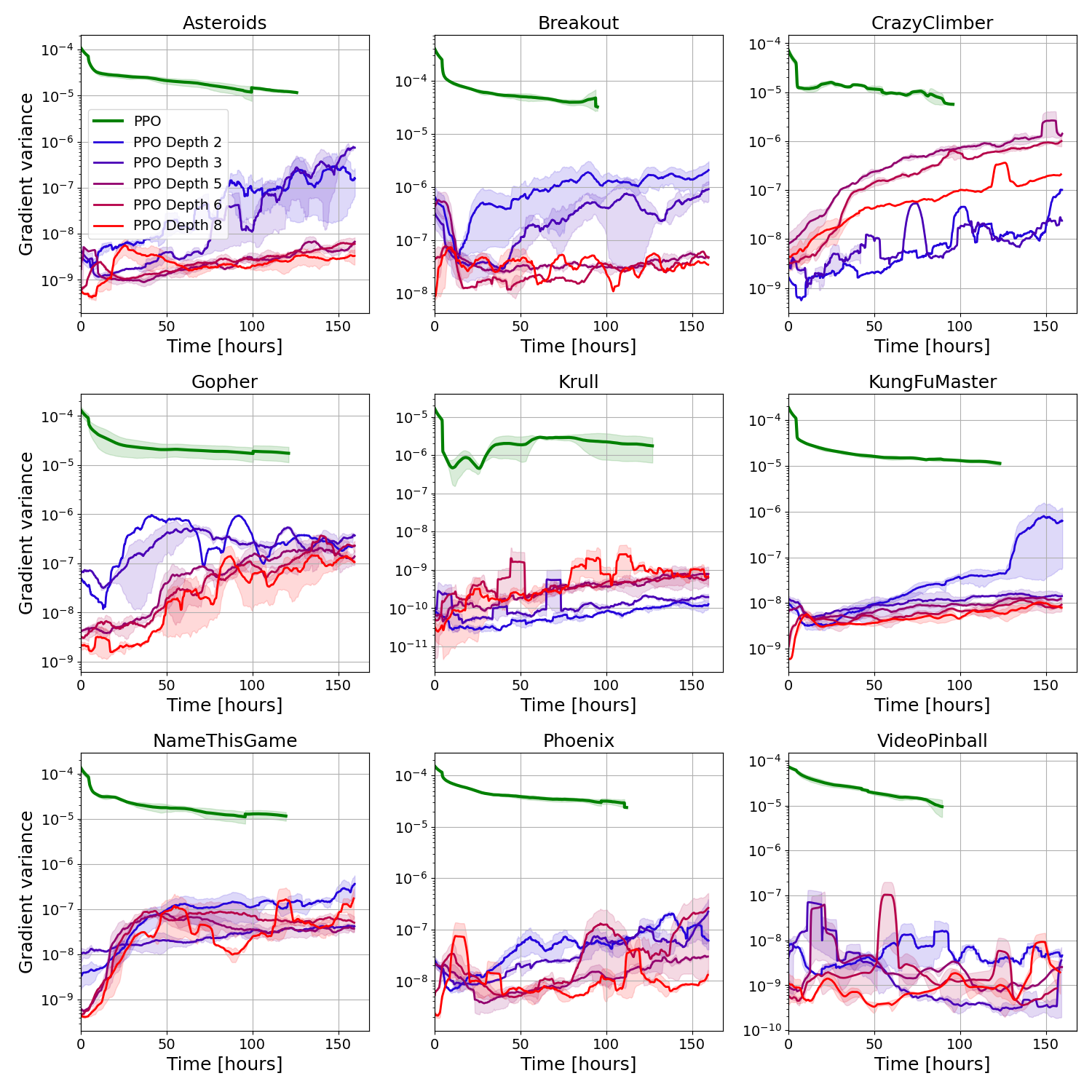}
    \caption{\textbf{Gradient variance: \treepol{} (single worker) vs PPO ($\bf{256}$ workers).} The results correspond to the same runs as in Figure~\ref{fig:train_curves}. Note the y-axis is in log-scale.}  \label{fig:variance_curves}
\end{figure}

In Figure~\ref{fig:variance_curves}, we show the gradient variance during training, as computed for all samples in the batch during the PPO training phase. As seen, the gradient variance of \treepol{} is consistently lower by three orders of magnitude than in PPO. We also observe a correlation between the gradient variance and the TS depth -- in most games, the variance monotonically reduces as the depth increases.


\section{Discussion}
Planning in RL is typically conducted with value-based algorithms due its seamless integration with the bellman operator, leaving aside the popular class of policy gradient methods. In this work, we introduced for the first time a tree-search approach for policy gradient. We show why our \treepol{} is essentially a variance reduction technique. Mitigating the known sample-inefficiency issue, it achieves better performance than the widely used PPO with multiple workers and soft-max policy.  Our method can be further applied to continuous control tasks, or in tasks where the forward model is learned with some estimation error. Other possible future directions are to study the theoretical implications of \treepol{} on the convergence rate of  policy gradient, or to extend it to adaptively changing depths.
\newpage
\bibliography{iclr2023_conference}
\bibliographystyle{iclr2023_conference}


\end{document}